\documentclass[runningheads]{llncs}
\usepackage{graphicx}
\usepackage{amsmath,amssymb} 
\usepackage{color}
\usepackage[ruled,vlined]{algorithm2e}
\usepackage{epsfig}
\usepackage{footnote}
\makesavenoteenv{tabular}
\makesavenoteenv{table}
\begin{document}
\title{C-WSL: Count-guided Weakly Supervised Localization} 

\titlerunning{C-WSL}
%
\author{Mingfei Gao\inst{1} \and Ang Li\inst{2}\thanks{The work was done while the author was at the University of Maryland} \and Ruichi Yu\inst{1} \and Vlad I. Morariu\inst{3\star} \and Larry S. Davis\inst{1}}
%
\authorrunning{M. Gao, A. Li, R. Yu, V. I. Morariu and L. S. Davis }
%

\institute{$^1$University of Maryland, College Park~~~$^2$DeepMind~~~~$^3$Adobe Research\\
	\email{ \{mgao,richyu,lsd\}@umiacs.umd.edu~~anglili@google.com~~morariu@adobe.com}
}
\maketitle              
\begin{abstract}
We introduce count-guided weakly supervised localization (C-WSL), an approach that uses per-class object count as a 
new form of supervision to improve weakly supervised localization (WSL). C-WSL uses a simple count-based region 
selection algorithm to select high-quality regions, each of which covers a single object instance during training, 
and improves existing WSL methods by training with the selected regions. To demonstrate the effectiveness of C-WSL, 
we integrate it into two WSL architectures and conduct extensive experiments on VOC2007 and VOC2012. Experimental 
results show that C-WSL leads to large improvements in WSL and that the proposed approach significantly outperforms 
the state-of-the-art methods. The results of annotation experiments on VOC2007 suggest that a modest extra time is 
needed to obtain per-class object counts compared to labeling only object categories in an image. Furthermore, we 
reduce the annotation time by more than $2\times$ and $38\times$ compared to center-click and bounding-box 
annotations.

\end{abstract}
\section{Introduction}
\label{sec: intro}

Convolutional neural networks (CNN) have achieved state-of-the-art performance on the object detection task~\cite{ren2015faster,liu2016ssd,redmon2016you,redmon2016yolo9000,singh2018analysis,lin2017focal,gao2018dynamic,lin2017feature,yang2016exploit,singh2018r,yu16objectcontextselection,yu2017_vrd_knowledge_distillation}. However, these detectors are trained in a strongly supervised setting, requiring a large number of bounding box annotations and huge amounts of human labor.
\begin{figure}
\centering
\includegraphics[width=1.0\linewidth]{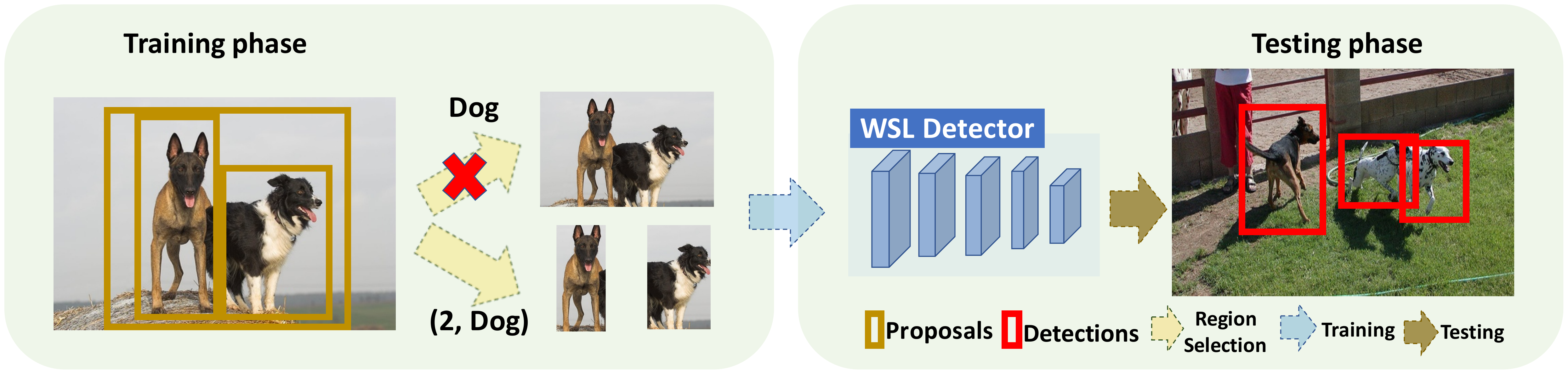}
\caption{Given a set of object proposals and the per-class object count label, we select high-quality positive regions (that tightly cover a single object) to train a WSL detector. Count information significantly reduces detected bounding boxes that are loose and contain two or more object instances, one of the most common errors produced by weakly supervised detectors}
\label{fig: idea}
\end{figure}

To ease the burden of human annotation, weakly supervised localization (WSL) methods train a detector using weak supervision, \emph{e.g.}, image-level supervision, instead of tight object bounding boxes. The presence of an object category in an image can be obtained on the Internet nearly for free, so most existing WSL architectures require only object categories as supervision. 

Existing methods~\cite{Bilen_2016_CVPR,cinbis2017weakly,diba2017weakly,kantorov2016contextlocnet,oquab2015object,wang2014weakly,tang2017multiple,Li_2016_CVPR,jie2017deep,zhu2017soft,kim2017two,shi2017weakly,singh2017hide} have proposed different architectures to address the WSL problem. However, there is still a large performance gap between weakly and strongly supervised detectors~\cite{ren2015faster,redmon2016yolo9000,liu2016ssd} on standard object detection benchmarks~\cite{pascal-voc-2007,pascal-voc-2012,lin2014microsoft}. Often, this is due to the limited information provided by object-category supervision. One major unsolved problem of WSL is that high confidence detections tend to include multiple objects instead of one. As shown in Fig.~\ref{fig: idea} (red cross branch), since training images containing multiple dogs are labeled just as ``Dog", detectors tend to learn the composite appearance of multiple dogs as if they were one dog and group multiple dogs as a single instance at test time. To resolve this ambiguity, we use per-class object count information to supervise detector training. 

Object count is a type of image-level supervision which is much weaker and cheaper than instance-level supervisions, such as center clicks~\cite{papadopoulos2017training} and bounding boxes. Unlike center click and bounding box annotations, which require several well-trained annotators to specify the center and tight box of each object, object count contains no location information and can be obtained without actually clicking on an object. Moreover, a widely studied phenomenon in psychology, called subitizing~\cite{clements1999subitizing} suggests that humans are able to determine the number of objects without pointing to or fixating on each object sequentially if the total number of objects in the image is small (typically 1-4)~\cite{chattopadhyay2017counting}. Thus, people may be able to specify the object count with just a glance. To demonstrate the inexpensiveness of count annotation, we conduct annotation experiments on Pascal VOC2007. Experimental results show that only a small amount of extra time is needed to obtain per-class object counts compared to labeling just object categories in an image and the response time of the count annotation is much less than that of object center and bounding box.

Our proposed method, Count-guided WSL (C-WSL), is illustrated in Fig.~\ref{fig: idea}. During the training process, C-WSL makes use of per-class object count supervision to identify the correct high-scoring object bounding boxes from a set of object proposals. Then, a weakly supervised detector is refined with these high-quality regions as pseudo ground-truth (GT) bounding boxes. This strategy is similar to existing WSL methods that refine detectors using automatically identified bounding boxes~\cite{Li_2016_CVPR,jie2017deep,tang2017multiple}. However, since these methods do not make use of object count supervision, they treat only the top-scoring region as the pseudo GT box, regardless of the number of object instances present in the image. This sometimes leads to multiple object instances being grouped into a single pseudo GT box, which hurts the detector's ability to localize individual objects. With the guidance of the object count label, C-WSL selects tight box regions that cover individual objects as shown in Fig.~\ref{fig: idea} (the ``(2, Dog)" branch).

The main contribution of C-WSL is that it uses per-class object count, a cheap and effective form of image-level supervision, to address a common failure case in WSL where one detected bounding box contains multiple object instances. To implement C-WSL, we develop a simple Count-based Region Selection (CRS) algorithm and integrate it into two existing architectures---alternating detector refinement (ADR) and online detector refinement (ODR)---to significantly improve WSL. Experimental results on Pascal VOC2007~\cite{pascal-voc-2007} and VOC2012~\cite{pascal-voc-2012} show that C-WSL significantly improves WSL detection and outperforms state-of-the-art methods.

\section{Related Works}
\label{sec: related}
\textit{MIL-based CNN Methods.} Most existing WSL methods~\cite{Bilen_2016_CVPR,cinbis2017weakly,diba2017weakly,kantorov2016contextlocnet,oquab2015object,wang2014weakly,tang2017multiple,Li_2016_CVPR,jie2017deep} are based on multiple instance learning (MIL)~\cite{dietterich1997solving}. In the MIL setting, a bag is defined as a collection of regions within an image. A bag is labeled as positive if at least one instance in the bag is positive and labeled as negative if all of its samples are negative.
Bilen~\emph{et~al.}~\cite{Bilen_2016_CVPR} proposed a two-stream CNN architecture to classify and localize simultaneously and train the network in an end-to-end manner. Following~\cite{Bilen_2016_CVPR}, Kantorov~\emph{et~al.}~\cite{kantorov2016contextlocnet} added \emph{additive} and \emph{contrastive} models to improve localization on object boundaries instead of local parts. Singh~\emph{et~al.}~\cite{singh2017hide} proposed the `Hide-and-Seek' framework which hides informative patches to encourage WSL to detect complete object instances. In~\cite{Li_2016_CVPR}, Li~\emph{et~al.} conducted progressive domain adaption and significantly improved the localization ability of the baseline detector. Diba~\emph{et~al.}~\cite{diba2017weakly} performed WSL in two/three cascaded stages to find the best candidate location based on a generated class activation map. Jie~\emph{et~al.} proposed a self-taught learning approach in~\cite{jie2017deep} which alternates between classifier training and online supportive sample harvesting. Similarly, in~\cite{tang2017multiple}, Tang~\emph{et~al.} designed an online classifier refinement pipeline to progressively locate the most discriminative region of an image. ~\cite{jie2017deep} and ~\cite{tang2017multiple} are most related to our approach since we also conduct alternating and online detector refinement. However, instead of using the top-scoring detection as the positive label~\cite{tang2017multiple} or mining confident regions by solving a complex dense subgraph discovery problem~\cite{jie2017deep}, we use per-class object count, a cheap form of supervision, to guide region selection and progressively obtain better positive training regions.
\begin{figure}[t]
\centering
   \includegraphics[width=0.8\linewidth]{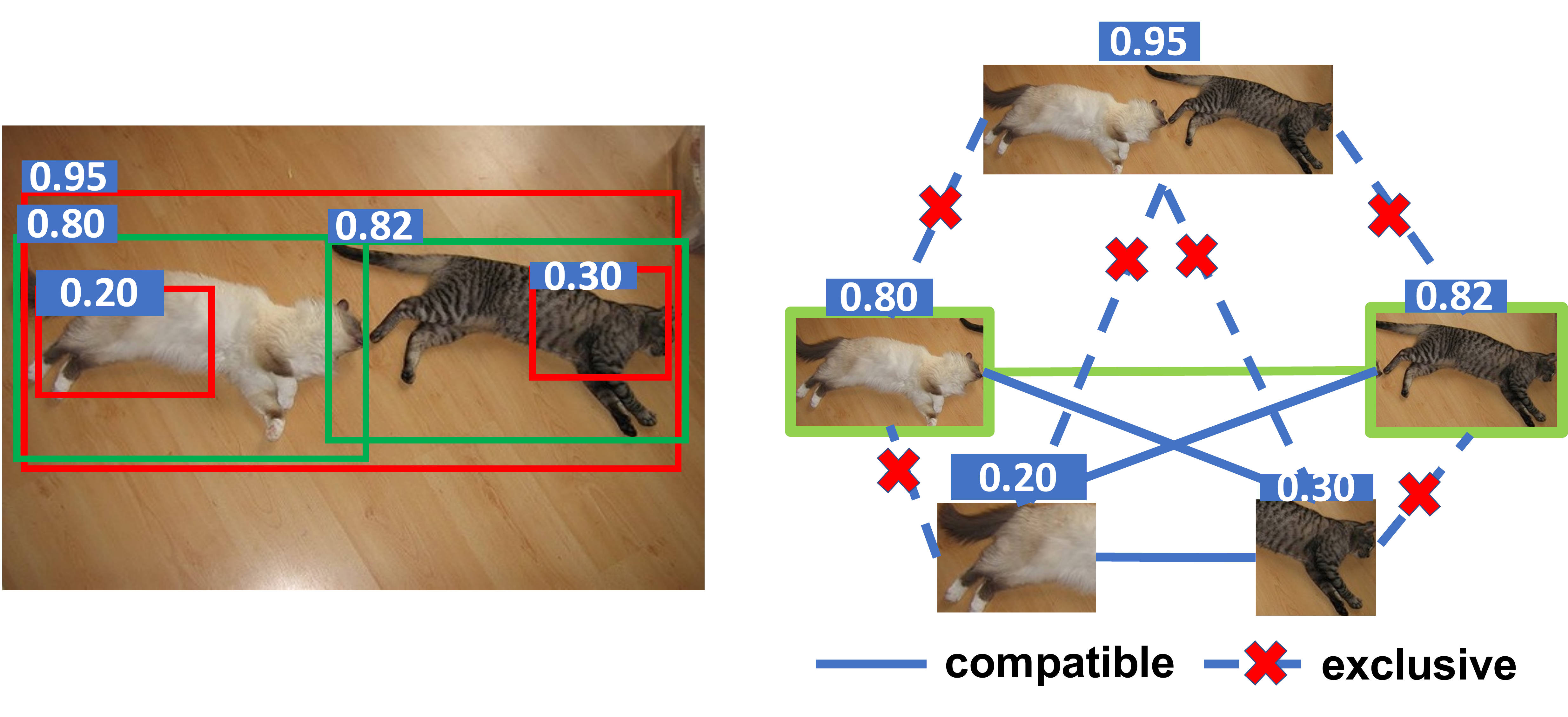}
   \caption{A common failure case of WSL methods (left) and graph representation of our region selection formulation (right). Our goal is to select the two green boxes, each of which tightly covers one object, as the positive training samples for WSL detectors. We achieve this by analyzing the confidence scores and spatial constraints among regions}
\label{fig: count_select}
\end{figure}

\textit{WSL with Different Supervisions.} \cite{papadopoulos2016we} proposed a novel framework where an annotator verifies predicted results instead of manually drawing boxes. Kolesnikov \emph{et~al.}~\cite{kolesnikov2016improving} assigned object or distractor labels to co-occuring objects in images to improve WSL. Papadopoulos~\emph{et~al.}~\cite{papadopoulos2017training} proposed click supervision and integrated it into existing MIL-based methods to improve localization performance. However, these methods either highly depend on the produced results and require frequent interactions with annotators or require annotators to search for and click on each instance in an image. In contrast, object count is an image-level annotation which contains no location information at all. It can be obtained with no clicks and few interactions, thus requires much less annotation time.

\section{Proposed Approach}
\label{sec: approaches}
C-WSL selects regions covering a single object with the help of per-class object count supervision and then refines the WSL detector using these regions as the pesudo GT bounding boxes. We first introduce a simple Count-based Region Selection (CRS) algorithm that C-WSL relies on to select high-quality regions from object proposals on training images. Then, we integrate CRS into two detector refinement structures to improve weakly supervised detectors.

\subsection{Count-based Region Selection (CRS)}
\label{sec: CRS}
As shown in Fig.~\ref{fig: count_select} (left), without object count information, previous methods often select the top-scoring box in training images as the positive training sample to refine the WSL detector~\cite{tang2017multiple,Li_2016_CVPR,jie2017deep}. Their detection performance is degraded because in many cases the top-scoring box contains multiple objects from the same category, \emph{e.g.}, two cats. Our goal is to select distinct regions, each covering a single object as positive training samples with the help of object count constraints so that the detector will learn the appearance of a single cat.

We formulate the problem as a region selection problem. Given a set of boxes $\mathbf{B}=\{b_1,...,b_N\}$ and the corresponding confidence scores $\mathbf{P}=\{p_1,...,p_N\}$ (\emph{e.g.}, the detection score of a region in each detector refinement iteration), a subset $\mathbf{G}$ is selected as the set of positive training regions where $|\mathbf{G}|=C$ and $C$ indicates the per-class object count. We identify a good subset $\mathbf{G}$ using a greedy algorithm applied to a graphical representation of the set of boxes. Each box is represented as a node in the graph, and two nodes are connected if the spatial overlap of their corresponding boxes is below a threshold (See solid line in Fig.~\ref{fig: count_select}). The greedy algorithm provides an approximation to the following optimization problem: 
\begin{equation}
\label{eq: select}
\begin{aligned}
&\mathbf{G^*} = \arg\max_\mathbf{G}\ {\sum_{b_k \in \mathbf{G}} {p_k}},&&\\
&s.t. \ |\mathbf{G}| = C, \ a_o(b_i, b_j) < T\ \forall b_i,b_j \in \mathbf{G},i\neq j.&&
\end{aligned}
\end{equation}
To encourage selecting regions containing just one object, we use the asymmetric area of overlap, i.e, $a_o(b_i, b_j) = \frac{area(b_i \cap b_j)}{area(b_j)}$, which has been proposed in~\cite{dollar2009integral,dollar2012pedestrian} to model spatial overlap between two boxes, where $b_i$ is a box previously selected by the greedy algorithm and $b_j$ indicates a box considered for selection. $T$ is the overlap threshold. If the algorithm has previously added a large box to the solution, thresholding on $a_o$ will discourage the selection of its subregions, regardless of their sizes.\footnote{The commonly used symmetric intersection-over-union measure would select sufficiently small regions even if they were fully overlapped by an existing large box.} So, to deliver a high total score, the algorithm prefers $C$ small high-scoring boxes to one large box, even though the large box may have the highest score. 

We conduct region selection after applying non-maximum suppression on a complete set of the detection boxes, so the number of nodes is limited to a reasonable number, and the computation cost is low in practice. The algorithm is summarized in~Alg.~\ref{alg: crs}.

\begin{algorithm}
\label{alg: crs}
\small
\SetAlgoVlined
\SetKwInOut{Input}{Input}
\SetKwInOut{Output}{Output}
\SetKwInput{Initialization}{Initialization}
\Input{$\mathbf{B} = \{b_1,...,b_N\}$, $\mathbf{P} = \{p_1,...,p_N\}$, $T$, $C$;\\ $\mathbf{B}$ is a list of candidate boxes; \\
$\mathbf{P}$ is the corresponding scores; \\
$T$ is the overlap threshold;  \\
$C$ indicates the object count;}
\Initialization{Sort (descend) $\mathbf{B}$ based on $\mathbf{P}$; \\$\mathbf{G^*} \leftarrow \emptyset$;  $s_{max} \leftarrow 0$;}
\Output{$\mathbf{G^*}$}
 \For{$i\in\{1,...,N\}$}{
 $\mathbf{G} \leftarrow b_i$; $s \leftarrow p_i$; \\
 \For{$j\in\{i+1,...,N\}$}{
  \If{\emph{$a_o(b_k,b_j)<T (\forall b_k \in  \mathbf{G})$}}{
   $\mathbf{G} \leftarrow \mathbf{G} \cup \{b_j\}$; $s \leftarrow s + p_j$ \\
   \If{$|\mathbf{G}|==C \ \text{or} \ j == N$}{
      \If{$s>s_{max}$}{
      $s_{max} \leftarrow s$; $\mathbf{G^*} \leftarrow
      \mathbf{G}$ \\}
      break;
      }
   }
  }
}
 \caption{Count-based Region Selection (CRS)}
\end{algorithm}

\begin{figure}[t]
\centering
  \includegraphics[width=1.0\linewidth]{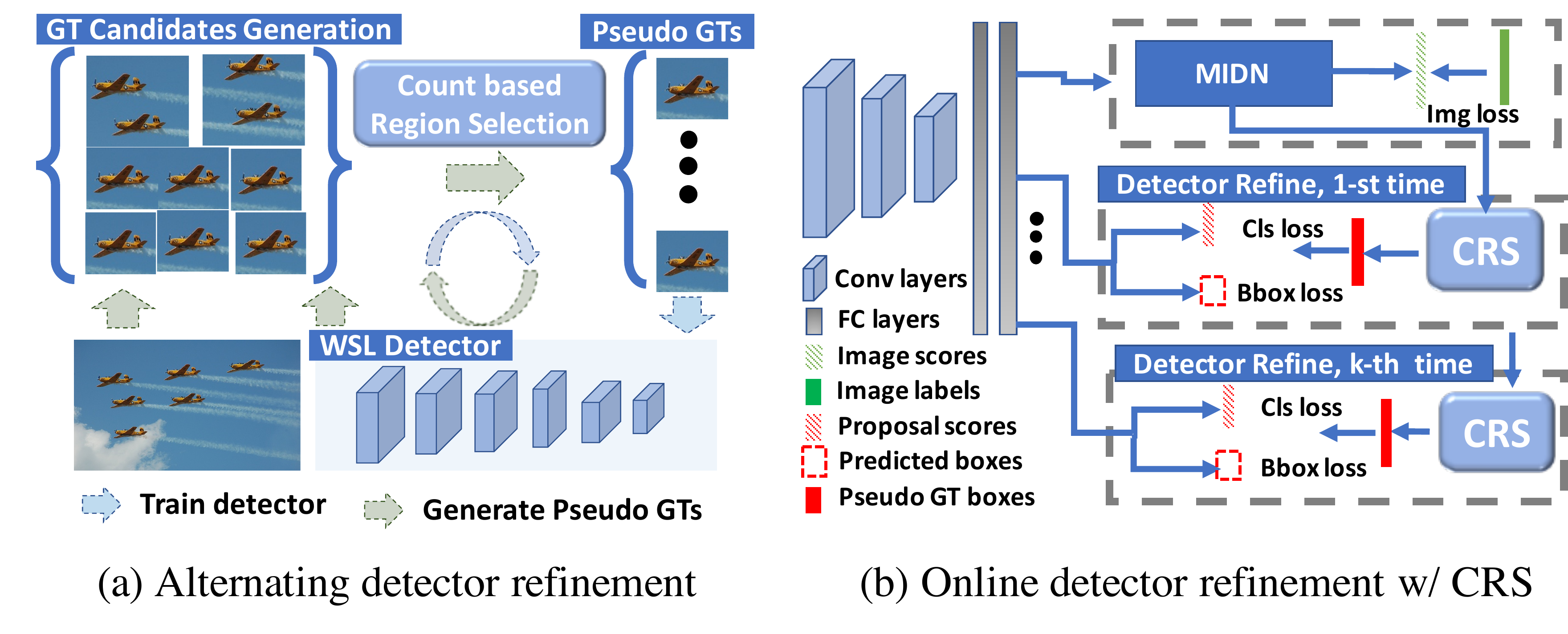}
   \caption{(a): Count-based Region Selection (\textit{CRS}) is applied to select high-quality positive training regions from the ground-truth (GT) candidate boxes generated by a WSL detector. The WSL detector is then refined using these regions. (b): The Multiple Instance Detection Network(\textit{MIDN})~\cite{Bilen_2016_CVPR,tang2017multiple} and multiple detector networks share the same feature representation to refine the detector at all stages together. \textit{Cls loss} indicates the classification loss and \textit{Bbox loss} indicates bounding box regression loss}
\label{fig: adr_odr}
\end{figure}
\subsection{Detector Refinement Structures with CRS}
\subsubsection{Alternating Detector Refinement (ADR).}
\label{sec: ADR}
We first integrate CRS into an alternating WSL refinement architecture, where a poor weakly supervised detector can be refined iteratively. The architecture is shown in Fig.~\ref{fig: adr_odr}, where a WSL detector alternates between generating high-quality regions as pseudo ground-truth (GT) boxes and refining itself using these GT boxes. Some WSL methods are based on a strategy like this~\cite{cinbis2017weakly,jie2017deep}. The major difference is that we use CRS to select multiple high-quality regions as the GT boxes.

\textit{Initialization phase.} We first generate a set of box candidates from the training data using a pre-trained WSL detector. This set of box candidates is treated as the initialized pseudo GTs and will be refined iteratively afterwards.

\textit{Alternating training phase.} We use Fast R-CNN~\cite{girshick2015fast} as our WSL network. Starting from the initialized pseudo GT boxes, Fast R-CNN alternates between improving itself via retraining with the pseudo GT boxes generated by CRS and generating a refined set of GT candidate boxes on the training images.

\subsubsection{Online Detector Refinement (ODR).}
\label{sec: ODR}
As argued in~\cite{tang2017multiple}, the alternating strategy has two potential limitations: 1) it is time consuming to alternate between training on the fixed labels and generating labels by the trained model; 2) separating refinements into different iterations might harm performance since it hinders the procedure from sharing image representations across iterations. 

Based on~\cite{tang2017multiple}, we propose an online detector refinement framework integrated with CRS. An illustration of the proposed method is shown in Fig.~\ref{fig: adr_odr}. A Multiple Instance Detection Network (MIDN) and several detector refinement stages share the same feature representation extracted from a backbone structure. The MIDN utilizes an object-category label to supervise its training as in~\cite{tang2017multiple,Bilen_2016_CVPR}. Each detector refinement network outputs the classification score and predicted bounding box for each region proposal. The predicted boxes with scores at each stage will be used to select pseudo GTs for the next stage refinement. Compared to~\cite{tang2017multiple}, we have two major differences: 1) we use CRS to generate high-quality regions as pseudo GTs rather than just choosing the top-scoring region; 2) we use both classification loss and bounding box regression loss for detector refinement, just as RCNNs do. Note that the inputs to CRS produced by MIDN are the proposals with scores before the summation over proposals.

\section{Experiments}
We compare with the existing WSL methods which are trained by object class labels to show the advantage of per-class count supervision. It may seem an `unfair' comparison, since the per-class count provides more information compared to object class. However, we demonstrate via our annotation experiment that the cost of the additional information is very low, which makes it reasonable to determine how much improvement can be gained by adding this information.
\subsection{Experimental Setup}
\textit{Datasets and Evaluate Metrics.} Comparisons with state-of-the-art methods are conducted on VOC2007~\cite{pascal-voc-2007} and VOC2012~\cite{pascal-voc-2012} which contain 20 object categories. For VOC2007, all the models are trained on the \emph{trainval} set which contains 5,011 images and evaluated on \emph{test} set which includes 4,952 images. For VOC2012, models are trained on 5,717 images of the \emph{train} set and evaluated on 5,823 images in the \emph{val} set. We use two widely used metrics for localization evaluation: Correct
localization (CorLoc)~\cite{oquab2015object} and Average Precision (AP)~\cite{everingham2010pascal}. CorLoc evaluates localization accuracy by measuring if the maximum response point of a detection is inside the ground truth bounding box. AP evaluates models by comparing \text{IoU} between output and ground truth bounding boxes. 

\textit{Implementation Details.}
We fix $T=0.1$ for all models at all the iterations on both datasets. Note that our experiments show that the method is robust to $T$, \emph{e.g.,} varying $T$ from 0.1 to 1 with step 0.1, we achieved (Mean, Std) = (47.2\%$, $ 0.42\%) mAP. Following~\cite{jie2017deep,tang2017multiple}, we set the total iteration number to 3 and use \emph{VGG16}~\cite{simonyan2014very} as the backbone structure for both ADR and ODR. For fair comparison, the existing works also use~\emph{VGG16} except for~\cite{cinbis2017weakly} which utilizes~\emph{AlexNet}.
In ADR, we strictly follow the steps of training Fast-RCNN at each iteration and use all the released default training parameters 
except that we use the generated pseudo GT boxes instead of the bounding box labels. In ODR, we follow the basic MIDN structure and training process from~\cite{tang2017multiple}, and use the parameters released by the author. 
Note that we use the same classification and bounding box regression loss in ODR as in~\cite{girshick2015fast}.

\textit{Variants of Our Approach.} \textit{C-WSL:WSLPDA/OICR+ADR} indicates ADR initialized with a pre-trained WSLPDA~\cite{Li_2016_CVPR} (or OICR~\cite{tang2017multiple}) model where CRS is used to select confident GT boxes in each iteration. Then, a Fast-RCNN is alternatively refined as we mentioned in Sec.~\ref{sec: ADR}. \textit{C-WSL:ODR} indicates the structure shown in Fig.~\ref{fig: adr_odr}(b). \textit{C-WSL:ODR+FRCNN} denotes a Fast RCNN trained with the top-scoring region generated by \textit{C-WSL:ODR} to improve results (inspired by~\cite{Li_2016_CVPR,tang2017multiple}). \textit{C-WSL*} indicates models trained by our annotated counts. 
\subsection{Annotation Time vs. Detection Accuracy}
\label{sec: time}
Object counting is very straightforward. The user interface includes an image and 15 buttons
indicating the count numbers. We cap object count with 15 since it is very rare to have a count of the same class bigger than 15. Similar to the click experiments [21], an annotator was given a category and was asked to click the count corresponding to that category. Following~\cite{papadopoulos2017training}, given an object category, we measure the response time of counting the object instances from the moment the image appears until the count is determined. 

 Annotation evaluations are conducted on the full \emph{trainval} set with 20 categories of VOC2007~\cite{pascal-voc-2007}. The average response time of counting a single object per class per image is $0.90s$. Average response time per image of annotating a single image class is from $1.5s$ to $1.9s$~\cite{krishna2016embracing} and that of annotating count given object class is $1.48s$, so obtaining per-class object count from an image only needs $1.48/1.9 = 78\%$ to $1.48/1.5 = 99\%$ more time compared to annotating just the object class.

\begin{figure}[t]
\centering
   \includegraphics[width=1.0\linewidth]{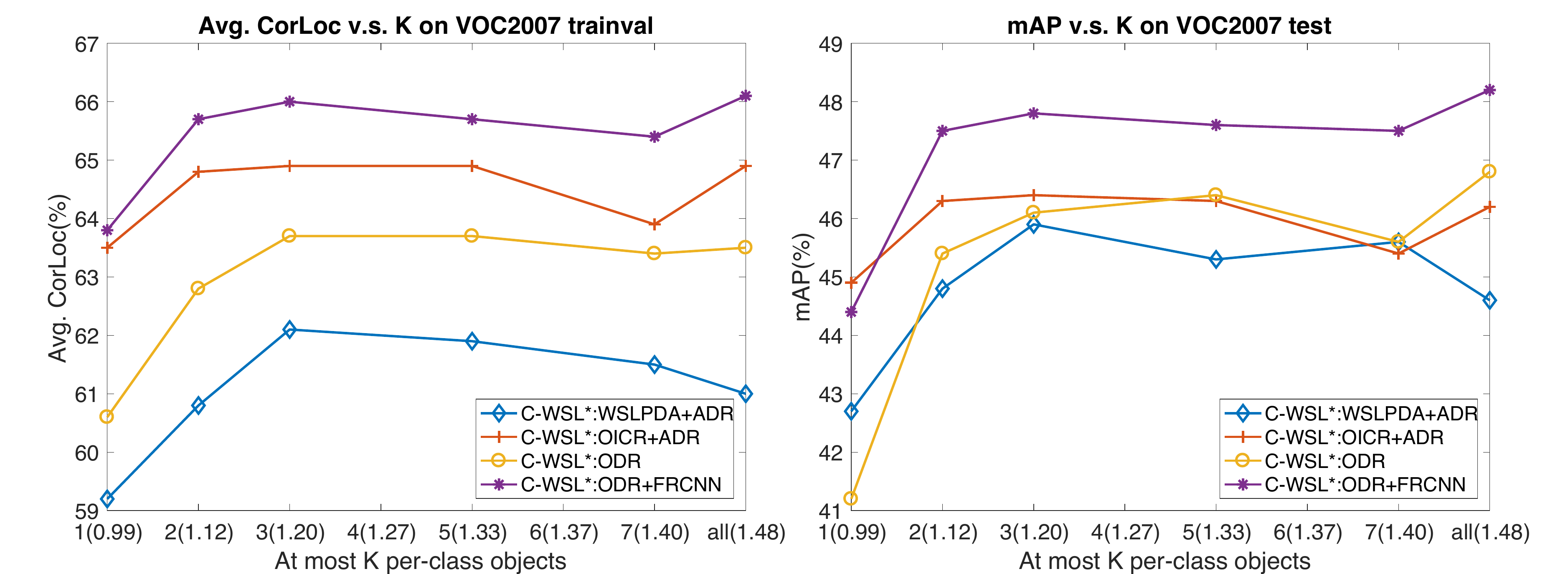}
   \caption{Detection accuracy analysis when at most \textit{K} per-class objects are counted in an image. Average annotation time (in seconds) per image under each \textit{K} is shown in the parentheses. Detection accuracy becomes stable when \textit{K}=3}
\label{fig: atmost}
\end{figure}

Annotation time of object counts per image increases as the number of objects increases. However, it might not always be helpful to count all the objects, especially for images with many objects, since these images are more likely to depict complex scenes, \emph{e.g.}, significant occlusions and small object instances, and for such images the generated GT candidates might not include all the objects in the first place. Thus, we evaluate the detection accuracy of our model using at most $K$ per-class objects annotation, where $K$ is the upper bound of per-class object instances that are counted for each image. Obviously, $K$ has positive correlation with annotation time, since annotators may not be able to subitize for high values of K and will need to spend an amount of time proportional to K in order to produce an accurate count. Analysis of \emph{mAP} and average \emph{CorLoc} vs. $K$ is shown in Fig.~\ref{fig: atmost}. The results suggest that the detection accuracy reaches the highest point when at most $3$ per-class objects are counted per image. Average annotation time per image for images with at most $3$ per-class objects is $1.20s$ which is $63\%\sim80\%$ overhead compared to object category annotations.
\begin{table}
\centering
\caption{Accuracy vs. cost among bounding box, clicks and count supervisions on VOC2007. We use~\cite{ren2015faster} as a reference of fully supervised detector}
\resizebox{\textwidth}{!}{%
\begin{tabular}{|c|c|c|c|c|}
\hline
Method & Faster-RCNN~\cite{ren2015faster} & Two-clicks~\cite{papadopoulos2017training} & One-click~\cite{papadopoulos2017training} & C-WSL*:ODR+FRCNN \\ \hline
mAP(\%) & 69.9 & 49.1(AlexNet)/57.5(VGG16) & 45.9(AlexNet) & 48.2(VGG16) \\ \hline
Annotation cost & \begin{tabular}[c]{@{}c@{}}34.5s/img+anno. train\\ +re-draw rejected boxes\end{tabular} & \begin{tabular}[c]{@{}c@{}}3.74s/img+anno. train\\ +re-click rejected clicks\end{tabular} & \begin{tabular}[c]{@{}c@{}}1.87s/img+anno. train\\ +re-click rejected clicks\end{tabular} & 0.90s/img \\ \hline
\end{tabular}
}
\label{tab: tradeoff}
\end{table}
We compare our models trained by our annotated counts and those obtained from the VOC2007 annotations in Tab.~\ref{tab: map} and ~\ref{tab: corloc}. The results demonstrate that models trained by the two sets of annotations have comparable performance, which suggests that our annotation is as useful as the VOC2007 annotations. Thus, in the following analysis, we just use (\textit{C-WSL}) VOC2007 annotations.

Accuracy and cost comparisons among box, clicks and count supervisions are shown in Tab.~\ref{tab: tradeoff}. Although the accuracy of our approach does not outperform supervised and two-click methods, we have achieved a
significant reduction in annotation cost. We are 38$\times$ and 4$\times$ faster regarding to response time for labeling a single image. In addition, box and clicks annotations require additional repeated annotator training to accurately locate objects and lengthy quality control processes. Our annotation does not require knowing the location of an object so it avoids the sensitivity to location noise. Consequently, we do not need annotator training and quality control in our experiments.

\subsection{Comparison with State-of-the-art (SOTA) Approaches}
\label{sec: comp_arts}
Comparison in terms of  \emph{mAP} on the VOC2007 \emph{test} set and \emph{CorLoc} on the VOC2007 \emph{trainval} set are shown in Tab.~\ref{tab: map} and~\ref{tab: corloc}, respectively. Overall, the proposed \emph{C-WSL:ODR+FRCNN} outperforms all the existing SOTA methods using both \emph{CorLoc} and \emph{mAP} measurements. 
\begin{table}
\setlength{\tabcolsep}{2pt}
\centering
\caption{Comparison with the state-of-the-art in terms of mAP on the VOC2007 \emph{test} set. Our number is marked in \textcolor{red}{red} if it is the best in the column}
\resizebox{\textwidth}{!}{%
\begin{tabular}{l|cccccccccccccccccccc|c}
\hline
Methods & are & bik & brd & boa & btl & bus & car & cat & cha & cow & tbl & dog & hrs & mbk & prs & plt & shp & sfa & trn & tv & mAP \\ \hline
Cinbis~\emph{et~al.}~\cite{cinbis2017weakly} &39.3 &43.0 &28.8 &20.4 &8.0 &45.5 &47.9 &22.1 &8.4 &33.5 &23.6 &29.2 &38.5 &47.9 &\textcolor{black}{20.3} &20.0 &35.8 &30.8 &41.0 &20.1 &30.2 \\
Wang~\emph{et~al.}~\cite{wang2014weakly} &48.8 &41.0 &23.6 &12.1 &11.1 &42.7 &40.9 &35.5 &11.1 &36.6 &18.4 &35.3 &34.8 &51.3 &17.2 &17.4 &26.8 &32.8 &35.1 &45.6 &30.9 \\
Jie~\emph{et~al.}~\cite{jie2017deep} & 52.2 &47.1 &35.0 &26.7 &15.4 &61.3 &66.0 &\textcolor{black}{54.3} &3.0 &53.6 &24.7 &\textcolor{black}{43.6} &48.4 &\textcolor{black}{65.8} &6.6 &18.8 &51.9 &43.6 &53.6 &62.4 &41.7\\ 
WSDDN~\cite{Bilen_2016_CVPR} & 39.4 &50.1 &31.5 &16.3 &12.6 &64.5 &42.8 &42.6 &10.1 &35.7 &24.9 &38.2 &34.4 &55.6 &9.4 &14.7 &30.2 &40.7 &54.7 &46.9 &34.8 \\ 
WSDDN+Context~\cite{kantorov2016contextlocnet} &  57.1 &52.0 &31.5 &7.6 &11.5 &55.0 &53.1 &34.1 &1.7 &33.1 &\textcolor{black}{49.2} &42.0 &47.3 &56.6 &15.3 &12.8 &24.8 &48.9 &44.4 &47.8 &36.3\\ 
WSDDN-Ens.~\cite{Bilen_2016_CVPR} &  46.4 &58.3 &35.5 &25.9 &14.0 &66.7 &53.0 &39.2 &8.9 &41.8 &26.6 &38.6 &44.7 &59.0 &10.8 &17.3 &40.7 &49.6 &56.9 &50.8 &39.3\\ 
WCCN-3stage~\cite{diba2017weakly} &49.5 &60.6 &38.6 &\textcolor{black}{29.2} &16.2 &70.8 &56.9 &42.5 &10.9 &44.1 &29.9 &42.2 &47.9 &64.1 &13.8 &23.5 &45.9 &54.1 &60.8 &54.5 &42.8 \\
WSLPDA~\cite{Li_2016_CVPR} & 54.5 &47.4 &41.3 &20.8 &17.7 &51.9 &63.5 &46.1 &21.8 &57.1 &22.1 &34.4 &50.5 &61.8 &16.2 &\textcolor{black}{29.9} &40.7 &15.9 &55.3 &40.2 &39.5 \\ 
OICR~\cite{tang2017multiple} &  58.0 &62.4 &31.1 &19.4 &13.0 &65.1 &62.2 &28.4 &24.8 &44.7 &30.6 &25.3 &37.8 &65.5 &15.7 &24.1 &41.7 &46.9 &64.3 &62.6 &41.2\\ 
OICR-Ens.+FRCNN\footnote{The numbers are reproduced by using the code released by the author.}~\cite{tang2017multiple} & \textcolor{black}{64.5}&\textcolor{black}{64.4}&44.1&25.9&16.9&67.8&68.4&33.2&9.0&57.5&46.4&21.7&\textcolor{black}{57.8}&64.3&10.0&23.7&50.6&60.9&\textcolor{black}{64.7}&58.0&45.5 \\
 \hline
\textbf{C-WSL:ODR} &\textbf{62.7}&\textbf{63.7}&\textbf{40.0}&\textbf{25.5}&\textcolor{black}{\textbf{17.7}}&\textbf{70.1}&\textbf{68.3}&\textbf{38.9}&\textbf{25.4}&\textbf{54.5}&\textbf{41.6}&\textbf{29.9}&\textbf{37.9}&\textbf{64.2}&\textbf{11.3}&\textcolor{black}{\textbf{27.4}}&\textbf{49.3}&\textbf{54.7}&\textbf{61.4}&\textcolor{red}{\textbf{67.4}}&\textbf{45.6}  \\ 
\textbf{C-WSL*:ODR} &\textcolor{black}{\textbf{62.9}}&\textcolor{black}{\textbf{64.8}}&\textbf{39.8}&\textbf{28.1}&\textbf{16.4}&\textbf{69.5}&\textbf{68.2}&\textbf{47.0}&\textcolor{black}{\textbf{27.9}}&\textbf{55.8}&\textbf{43.7}&\textbf{31.2}&\textbf{43.8}&\textbf{65.0}&\textbf{10.9}&\textbf{26.1}&\textbf{52.7}&\textbf{55.3}&\textbf{60.2}&\textcolor{black}{\textbf{66.6}}&\textbf{46.8} \\
\textbf{C-WSL:ODR+FRCNN} &\textbf{61.9}&\textbf{61.9}&\textcolor{black}{\textbf{48.6}}&\textcolor{black}{\textbf{28.7}}&\textcolor{red}{\textbf{23.3}}&\textbf{\textcolor{red}{71.1}}&\textcolor{red}{\textbf{71.3}}&\textbf{38.7}&\textbf{\textcolor{red}{28.5}}&\textcolor{red}{\textbf{60.6}}&\textbf{45.4}&\textbf{26.3}&\textbf{49.7}&\textbf{65.5}&\textbf{7.2}&\textbf{27.3}&\textcolor{black}{\textbf{54.7}}&\textcolor{red}{\textbf{61.6}}&\textbf{63.2}&\textbf{59.5}&\textbf{\textcolor{black}{47.8}}  \\
\textbf{C-WSL*:ODR+FRCNN} &\textcolor{black}{\textbf{62.9}}&\textcolor{red}{\textbf{68.3}}&\textcolor{red}{\textbf{52.9}}&\textbf{25.8}&\textbf{16.5}&\textcolor{red}{\textbf{71.1}}&\textcolor{black}{\textbf{69.5}}&\textcolor{black}{\textbf{48.2}}&\textbf{26.0}&\textcolor{black}{\textbf{58.6}}&\textbf{44.5}&\textbf{28.2}&\textbf{49.6}&\textcolor{red}{\textbf{66.4}}&\textbf{10.2}&\textbf{26.4}&\textcolor{red}{\textbf{55.3}}&\textbf{59.9}&\textbf{61.6}&\textbf{62.2}&\textbf{\textcolor{red}{48.2}}\\\hline
\end{tabular}
}
\label{tab: map}

\end{table}

\begin{table*}
\footnotesize
\setlength{\tabcolsep}{2pt}
\centering
\caption{Comparison with the state-of-the-art in terms of CorLoc ($\%$) on the VOC2007 \emph{trainval} set. Our number is marked in \textcolor{red}{red} if it is the best in the column}
\resizebox{\textwidth}{!}{%
\begin{tabular}{l|cccccccccccccccccccc|c}
\hline
Methods & are & bik & brd & boa & btl & bus & car & cat & cha & cow & tbl & dog & hrs & mbk & prs & plt & shp & sfa & trn & tv & Avg. \\ \hline
Cinbis~\emph{et~al.}~\cite{cinbis2017weakly} &65.3 &55.0 &52.4 &48.3 &18.2 &66.4 &77.8 &35.6 &26.5 &67.0 &46.9 &48.4 &70.5 &69.1 &35.2 &35.2 &69.6 &43.4 &64.6 &43.7 &52.0 \\
Wang~\emph{et~al.}~\cite{wang2014weakly} &80.1 &63.9 &51.5 &14.9 &21.0 &55.7 &74.2 &43.5 &26.2 &53.4 &16.3 &56.7 &58.3 &69.5 &14.1 &38.3 &58.8 &47.2 &49.1 &60.9 &48.5 \\
Jie~\emph{et~al.}~\cite{jie2017deep} & 72.7 & 55.3 & 53.0 & 27.8 & 35.2 & 68.6 & 81.9 & 60.7 & 11.6 & 71.6 & 29.7 & 54.3 & 64.3 & 88.2 & 22.2 & 53.7 & 72.2 & 52.6 & 68.9 & 75.5 & 56.1 \\ 
WSDDN~\cite{Bilen_2016_CVPR} & 65.1 & 58.8 & 58.5 & 33.1 & 39.8 & 68.3 & 60.2 & 59.6 & 34.8 & 64.5 & 30.5 & 43.0 & 56.8 & 82.4 & 25.5 & 41.6 & 61.5 & 55.9 & 65.9 & 63.7 & 53.5 \\ 
WSDDN+Context~\cite{kantorov2016contextlocnet} & 83.3 & 68.6 & 54.7 & 23.4 & 18.3 & 73.6 & 74.1 & 54.1 & 8.6 & 65.1 & 47.1 & 59.5 & 67.0 & 83.5 & 35.3 & 39.9 & 67.0 & 49.7 & 63.5 & 65.2 & 55.1 \\ 
WSDDN-Ens.~\cite{Bilen_2016_CVPR} & 68.9 & 68.7 & 65.2 & 42.5 & 40.6 & 72.6 & 75.2 & 53.7 & 29.7 & 68.1 & 33.5 & 45.6 & 65.9 & 86.1 & 27.5 & 44.9 & 76.0 & 62.4 & 66.3 & 66.8 & 58.0 \\ 
WCCN-3stage~\cite{diba2017weakly} &83.9 &72.8 &64.5 &44.1 &40.1 &65.7 &82.5 &58.9 &33.7 &72.5 &25.6 &53.7 &67.4 &77.4 &26.8 &49.1 &68.1 &27.9 &64.5 &55.7 &56.7\\
SP-VGGNet~\cite{zhu2017soft} &85.3 &64.2 &\textcolor{black}{67.0} &42.0 &16.4 &71.0 &64.7 &\textcolor{black}{88.7} &20.7 &63.8 &\textcolor{black}{58.0} &\textcolor{black}{84.1} &\textcolor{black}{84.7} &80.0 &\textcolor{black}{60.0} &29.4 &56.3 &68.1 &77.4 &30.5 &60.6\\
WSLPDA~\cite{Li_2016_CVPR} & 78.2 & 67.1 & 61.8 & 38.1 & 36.1 & 61.8 & 78.8 & 55.2 & 28.5 & 68.8 & 18.5 & 49.2 & 64.1 & 73.5 & 21.4 & 47.4 & 64.6 & 22.3 & 60.9 & 52.3 & 52.4 \\ 
OICR~\cite{tang2017multiple} & 81.7 & \textcolor{black}{80.4} & 48.7 & 49.5 & 32.8 & 81.7 & 85.4 & 40.1 & 40.6 & 79.5 & 35.7 & 33.7 & 60.5 & 88.8 & 21.8 & 57.9 & 76.3 & 59.9 & 75.3 & 81.4 & 60.6 \\ 
OICR-Ens.+FRCNN$^2$~\cite{tang2017multiple} &\textcolor{black}{88.3}&78.8&62.8&48.9&38.9&83.2&85.4&50.0&21.9&77.4&45.6&41.9&79.3&91.6&12.6&60.8&\textcolor{black}{86.6}&\textcolor{black}{70.2}&80.2&79.9&64.2  \\\hline
\textbf{C-WSL:ODR} &\textbf{86.3}&\textbf{80.4}&\textbf{58.3}&\textbf{50.0}&\textbf{36.6}&\textbf{\textcolor{red}{85.8}}&\textbf{86.2}&\textbf{47.1}&\textbf{42.7}&\textcolor{black}{\textbf{81.5}}&\textbf{42.2}&\textbf{42.6}&\textbf{50.7}&\textbf{90.0}&\textbf{14.3}&\textbf{\textcolor{red}{61.9}}&\textcolor{black}{\textbf{85.6}}&\textbf{64.2}&\textbf{77.2}&\textbf{82.4}&\textbf{63.3}   \\ 
\textbf{C-WSL*:ODR} &\textbf{85.8}&\textbf{\textcolor{black}{81.2}}&\textbf{64.9}&\textbf{\textcolor{black}{50.5}}&\textbf{32.1}&\textbf{\textcolor{black}{84.3}}&\textbf{85.9}&\textbf{54.7}&\textbf{43.4}&\textbf{80.1}&\textbf{42.2}&\textbf{42.6}&\textbf{60.5}&\textbf{90.4}&\textbf{13.7}&\textbf{57.5}&\textbf{82.5}&\textbf{61.8}&\textbf{74.1}&\textbf{82.4}&\textbf{63.5} \\ 
\textbf{C-WSL:ODR+FRCNN} &\textbf{85.8}&\textbf{78.0}&\textbf{61.6}&\textbf{\textcolor{red}{52.1}}&\textbf{\textcolor{red}{44.7}}&\textbf{81.7}&\textbf{\textcolor{red}{88.4}}&\textbf{49.1}&\textbf{\textcolor{black}{50.0}}&\textbf{\textcolor{red}{82.9}}&\textbf{44.1}&\textbf{44.4}&\textbf{63.9}&\textcolor{black}{\textbf{92.4}}&\textbf{14.3}&\textbf{60.4}&\textcolor{red}{\textbf{86.6}}&\textbf{68.3}&\textcolor{red}{\textbf{80.6}}&\textbf{\textcolor{black}{82.8}}&\textbf{\textcolor{black}{65.6}} \\
\textbf{C-WSL*:ODR+FRCNN} &\textbf{\textcolor{black}{87.5}}&\textbf{\textcolor{red}{81.6}}&\textbf{\textcolor{black}{65.5}}&\textbf{\textcolor{red}{52.1}}&\textbf{37.4}&\textbf{83.8}&\textbf{\textcolor{black}{87.9}}&\textbf{57.6}&\textbf{\textcolor{red}{50.3}}&\textbf{80.8}&\textbf{44.9}&\textbf{44.4}&\textbf{65.6}&\textcolor{red}{\textbf{92.8}}&\textbf{14.9}&\textbf{\textcolor{black}{61.2}}&\textbf{83.5}&\textcolor{black}{\textbf{68.5}}&\textbf{77.6}&\textbf{\textcolor{red}{83.5}}&\textbf{\textcolor{red}{66.1}} \\
\hline
\end{tabular}
}
\label{tab: corloc}

\end{table*}
\begin{table*}
\footnotesize
\setlength{\tabcolsep}{2pt}
\centering
\caption{Comparison with baselines in terms of mAP on the VOC2007 \emph{test} set. The table contains two comparison groups separated by double solid lines. Each group shows how much ADR and C-WSL improve each baseline. \underline{Underline} is used if the C-WSL variant outperforms its baselines}
\resizebox{\textwidth}{!}{%
\begin{tabular}{l|cccccccccccccccccccc|c}
\hline
Methods & are & bik & brd & boa & btl & bus & car & cat & cha & cow & tbl & dog & hrs & mbk & prs & plt & shp & sfa & trn & tv & mAP \\ \hline
WSLPDA~\cite{Li_2016_CVPR} & 54.5 &47.4 &41.3 &20.8 &17.7 &51.9 &63.5 &46.1 &21.8 &57.1 &22.1 &34.4 &50.5 &61.8 &16.2 &29.9 &40.7 &15.9 &55.3 &40.2 &39.5 \\ 
WSLPDA+ADR &57.9&68.3&47.8&20.3&12.2&52.9&67.6&68.8&24.6&50.0&24.9&49.8&54.8&63.5&14.1&27.4&41.2&19.5&57.1&30.7&42.7\\ 
\textbf{C-WSL:WSLPDA+ADR}&\underline{\textbf{60.5}}&\textbf{\underline{70.1}}&\textbf{\underline{52.5}}&\underline{\textbf{24.7}}&\textbf{\underline{24.4}}&\underline{\textbf{63.6}}&\underline{\textbf{71.8}}&\textbf{58.1}&\underline{\textbf{26.0}}&\textbf{\underline{66.4}}&\underline{\textbf{26.5}}&\textbf{34.7}&\underline{\textbf{55.0}}&\textbf{\underline{65.8}}&\textbf{8.8}&\textbf{\underline{31.9}}&\underline{\textbf{51.6}}&\underline{\textbf{20.4}}&\underline{\textbf{60.0}}&\underline{\textbf{41.8}}&\underline{\textbf{45.7}}   \\ 
\hline \hline
OICR~\cite{tang2017multiple} &  58.0 &62.4 &31.1 &19.4 &13.0 &65.1 &62.2 &28.4 &24.8 &44.7 &30.6 &25.3 &37.8 &65.5 &15.7 &24.1 &41.7 &46.9 &64.3 &62.6 &41.2\\ 
OICR+ADR &58.1&61.2&43.3&24.4&19.4&65.5&67.1&34.3&3.6&56.5&45.5&26.4&61.9&60.7&10.4&23.6&49.2&62.1&61.4&64.2&44.9\\
\textbf{C-WSL:OICR+ADR} &\underline{\textbf{61.7}}&\underline{\textbf{66.8}}&\underline{\textbf{45.6}}&\textbf{21.1}&\underline{\textbf{23.5}}&\underline{\textbf{67.2}}&\textbf{\underline{73.8}}&\textbf{32.5}&\textbf{10.6}&\textbf{54.6}&\textbf{42.9}&\textbf{16.6}&\textbf{59.2}&\textbf{63.3}&\textbf{11.0}&\underline{\textbf{25.4}}&\textbf{\underline{55.3}}&\textbf{61.3}&\textbf{\underline{67.4}}&\textbf{\underline{67.8}}&\underline{\textbf{46.4}}
  \\\hline
\end{tabular}
}
\label{tab: adr_map}

\end{table*}
\begin{table*}
\footnotesize
\setlength{\tabcolsep}{2pt}
\centering
\caption{Comparison with the baseline detectors in terms of CorLoc ($\%$) on the VOC2007 \emph{trainval} set. The table contains two comparison groups separated by double solid lines. Each group shows how much ADR and C-WSL improve each baseline. \underline{Underline} is used if the C-WSL variant outperforms its baselines}
\resizebox{\textwidth}{!}{%
\begin{tabular}{l|cccccccccccccccccccc|c}
\hline
Methods & are & bik & brd & boa & btl & bus & car & cat & cha & cow & tbl & dog & hrs & mbk & prs & plt & shp & sfa & trn & tv & Avg. \\ \hline
WSLPDA~\cite{Li_2016_CVPR} & 78.2 & 67.1 & 61.8 & 38.1 & 36.1 & 61.8 & 78.8 & 55.2 & 28.5 & 68.8 & 18.5 & 49.2 & 64.1 & 73.5 & 21.4 & 47.4 & 64.6 & 22.3 & 60.9 & 52.3 & 52.4 \\ 
WSLPDA+ADR & 84.6&76.9&69.7&41.0&21.8&68.5&83.2&77.6&34.4&76.7&19.8&73.7&75.2&84.7&26.3&53.8&70.1&22.3&73.8&50.9
&59.2   \\ 
\textbf{C-WSL:WSLPDA+ADR} &\textbf{83.3}&\underline{\textbf{80.0}}&\textbf{\underline{70.9}}&\underline{\textbf{51.6}}&\underline{\textbf{41.2}}&\underline{\textbf{73.6}}&\underline{\textbf{85.3}}&\textbf{67.7}&\underline{\textbf{40.7}}&\underline{\textbf{79.5}}&\underline{\textbf{20.9}}&\textbf{54.7}&\underline{\textbf{79.6}}&\underline{\textbf{87.1}}&\textbf{24.5}&\underline{\textbf{56.8}}&\underline{\textbf{83.5}}&\textbf{20.7}&\underline{\textbf{76.0}}&\underline{\textbf{60.2}}&\underline{\textbf{61.9}} \\ 
\hline \hline
OICR~\cite{tang2017multiple} & 81.7 & 80.4 & 48.7 & 49.5 & 32.8 & 81.7 & 85.4 & 40.1 & 40.6 & 79.5 & 35.7 & 33.7 & 60.5 & 88.8 & 21.8 & 57.9 & 76.3 & 59.9 & 75.3 & 81.4 & 60.6 \\ 
OICR+ADR &85.8&76.9&65.8&49.5&38.5&83.2&84.8&49.7&14.0&79.5&46.8&41.2&80.3&89.2&15.0&60.1&84.5&66.4&78.3&80.6&63.5 \\
\textbf{C-WSL:OICR+ADR} &\textbf{85.4}&\textbf{78.0}&\textbf{65.5}&\underline{\textbf{49.5}}&\underline{\textbf{43.5}}&\underline{\textbf{84.3}}&\underline{\textbf{87.5}}&\textbf{48.0}&\textbf{23.6}&\underline{\textbf{80.8}}&\textbf{43.3}&\textbf{38.8}&\textbf{79.9}&\textbf{\underline{92.8}}&\textbf{15.8}&\underline{\textbf{60.1}}&\textbf{\underline{87.6}}&\underline{\textbf{66.4}}&\textbf{\underline{81.0}}&\textbf{80.3}&\underline{\textbf{64.6}} \\
\hline
\end{tabular}
}
\label{tab: adr_corloc}
\end{table*}
\begin{figure}[t]
\centering
   \includegraphics[width=1.0\linewidth]{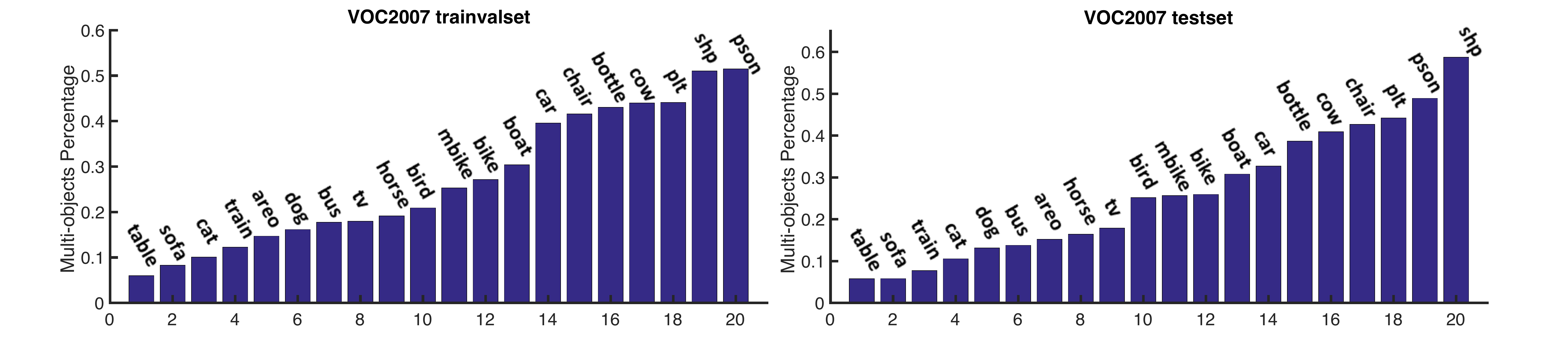}
   \caption{Image number of multiple-objects over image number of non-zero objects. Note that ``pson" means "person", ``plt" means "plant" and ``shp" denotes ``sheep". C-WSL works better on most classes with high multiple-objects percentage. See Sec.~\ref{sec: comp_arts}}
\label{fig: count_pclass}
\end{figure}
Tab.~\ref{tab: adr_map} and~\ref{tab: adr_corloc} compare our variants with the two baseline detectors, \emph{i.e.}, WSLPDA~\cite{Li_2016_CVPR} and OICR~\cite{tang2017multiple}. The results suggest that even the simple ADR strategy can significantly improve the results. Moreover, if we use object count information, we can largely improve WSLPDA by 6.2$\%$ \emph{mAP} (9.5$\%$ average \emph{CorLoc}) and OICR by 5.2$\%$ \emph{mAP} (4.0$\%$ average \emph{CorLoc}). C-WSL improves the results of \emph{WSLPDA+ADR} on 17 (15) out of 20 categories and the results of \emph{OICR+ADR} on 10 (10) out of 20 categories  in terms of \emph{mAP} on the VOC2007 \emph{test} set (in terms of \emph{CorLoc} on the VOC2007 \emph{trainval} set).

\begin{figure}
\centering
   \includegraphics[width=1.0\linewidth]{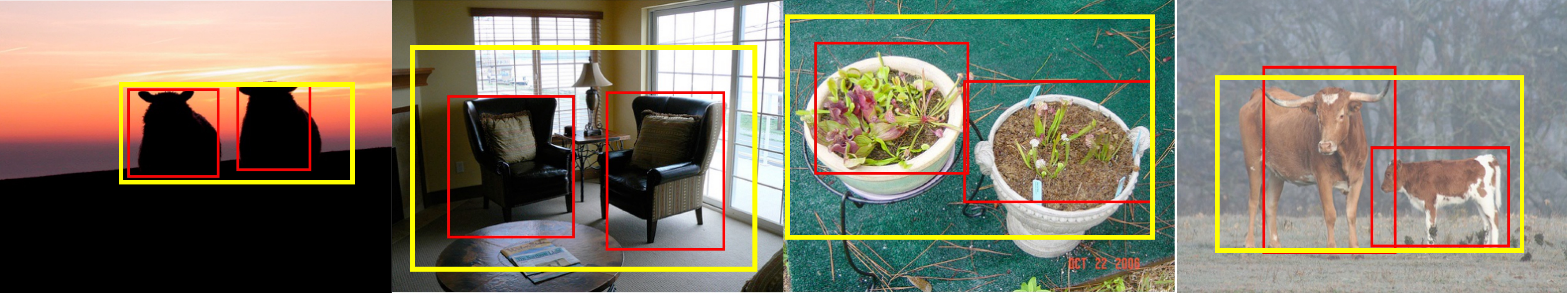}
   \caption{Examples of the training regions selected by \emph{OICR+CRS} (red) and \emph{OICR} (yellow). The regions selected by \emph{OICR} contain multiple object instances. Object count information helps to select regions, each covering a single instance}
\label{fig: train_region}
\end{figure}

\begin{figure*}
\centering
   \includegraphics[width=1.0\linewidth]{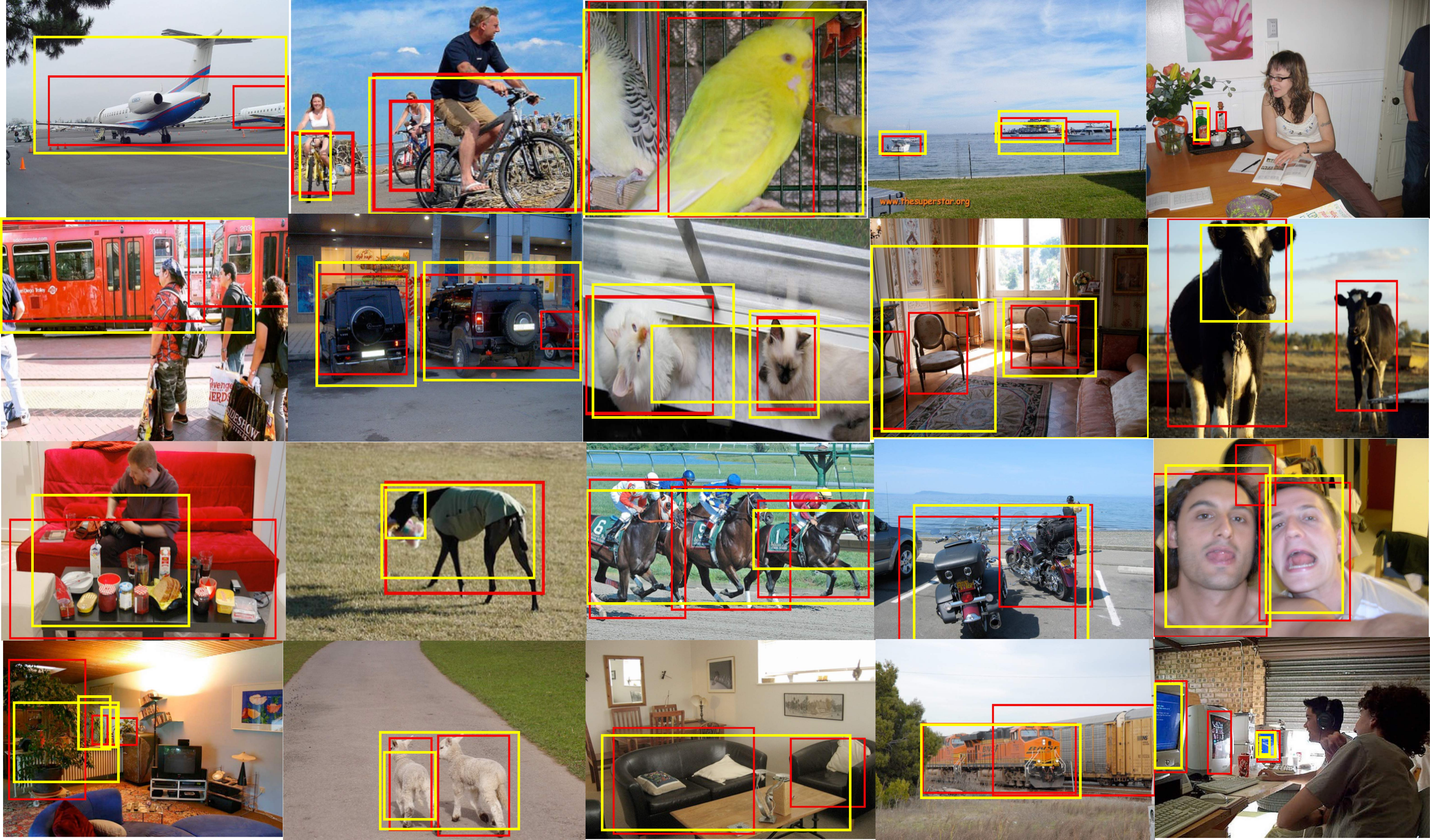}
   \caption{Qualitative comparison between our  \emph{CWSL:ODR+FRCNN} (red boxes) and \emph{OICR+FRCNN} (yellow boxes) on the VOC2007 \emph{test} set over the 20 classes. Our detector detects much tighter bounding boxes, yields much fewer boxes with multiple objects in them, and finds instances more accurately}
\label{fig: qualitative}
\end{figure*}

As stated in Sec.~\ref{sec: intro}, the object count information is helpful to avoid a detector localizing on multiple objects. To demonstrate this point, we first calculate the percentage of images that have more than one per-class object (multi-objects percentage) in VOC2007. As shown in Fig.~\ref{fig: count_pclass}, ``bottle", ``car", ``chair", ``cow", ``person", ``plant" and ``sheep" have a high percentage of images which include more than one object in the corresponding category. As shown in Tab.~\ref{tab: map} and~\ref{tab: corloc}, \emph{C-WSL:ODR+FRCNN} outperforms SOTA methods for 5 out of these 7 categories. When looking into the effect of object count supervision on WSLPDA and OICR, we see significant improvement on these categories as shown in Tab.~\ref{tab: adr_map} and~\ref{tab: adr_corloc}. Consider the ``sheep" category for example. \emph{C-WSL:WSLPDA+ADR} improves \emph{WSLPDA+ADR} by $13.4\%$ \emph{CorLoc} and $10.4\%$ \emph{AP}. \emph{C-WSL:OICR+ADR} improves \emph{OICR+ADR} by $3.1\%$ \emph{CorLoc} and $6.1\%$ \emph{AP}. Fig.~\ref{fig: train_region} shows some examples of training regions selected by \emph{OICR+CRS} and \emph{OICR}. \emph{OICR} tends to select regions containing multiple instances, while object count helps to obtain regions including a single instance. Qualitative comparison between our \emph{C-WSL:ODR+FRCNN} and \emph{OICR-Ens.+FRCNN} on the VOC2007 \emph{test} set is shown in Fig.~\ref{fig: qualitative}, demonstrating that our approach achieves more precise localization when multiple per-class objects appear in an image. We will further analyze our approach on images with different numbers of objects in Sec.~\ref{sec: abl-counts}.

Tab.~\ref{tab: map_2012} and~\ref{tab: corloc_2012} show the comparison of C-WSL with the SOTA on VOC2012. Note that results of WSLPDA and OICR models are reproduced by running the pretrained model and the code released by the authors. The results suggest that our method outperforms the SOTA method (\emph{OICR-Ens.+FRCNN}) by $2.6\%$ in \emph{mAP} on the VOC2012 \emph{val} set and by $2.8\%$ in \emph{CorLoc} on the VOC2012 \emph{train} set. C-WSL improves the results of \emph{WSLPDA+ADR} on 12 (10) out of 20 categories and the results of \emph{OICR+ADR} on 10 (12) out of 20 categories  in terms of \emph{mAP} on the VOC2012 \emph{val} set (in terms of \emph{CorLoc} on the VOC2012 \emph{train} set).
\begin{table*}
\footnotesize
\setlength{\tabcolsep}{2pt}
\centering
\caption{Comparison with the state-of-the-art in terms of mAP on the VOC2012 \emph{val} set. Our number is marked in \textcolor{red}{red} if it is the best in the column. \underline{Underline} is used if the C-WSL variant outperforms its baselines}
\resizebox{\textwidth}{!}{%
\begin{tabular}{l|cccccccccccccccccccc|c}
\hline
Methods & are & bik & brd & boa & btl & bus & car & cat & cha & cow & tbl & dog & hrs & mbk & prs & plt & shp & sfa & trn & tv & mAP \\ \hline
Jie~\emph{et~al.}~\cite{jie2017deep} &60.9& 53.3 &31.0 &16.4 &18.2 &58.2 &50.5 &55.6 &9.1 &42.1 &12.1 &\textcolor{black}{43.4} &45.3 &64.6 &7.4 &19.3 &44.8 &39.3 &51.4 &57.2 &39.0 \\
OICR-Ens.+FRCNN~\cite{tang2017multiple}&71.0&68.2&\textcolor{black}{52.7}&20.1&27.2&57.3&\textcolor{black}{57.1}&19.0&8.0&50.6&\textcolor{black}{30.2}&34.5&63.3&69.5&1.2&20.5&48.5&\textcolor{black}{55.2}&41.1&60.4&42.8 \\
\hline
WSLPDA~\cite{Li_2016_CVPR} & 42.2 &27.8 &32.7 &4.2 &13.7 &52.1 &35.8 &48.3 &11.8 &31.7 &4.9 &30.4 &45.3 &51.8 &\textcolor{black}{11.5} &13.4 &33.5 &7.2 &45.6 &38.4 &29.1 \\ 
WSLPDA+ADR &70.0&65.6&46.3&14.4&22.8&57.5&54.2&\textcolor{black}{67.5}
&16.1&45.0&4.4&40.0&51.7&71.8&5.8&\textcolor{black}{27.7}&38.3&11.7&55.2&34.1&40.0\\ 
\textbf{C-WSL:WSLPDA+ADR}&\textbf{69.8}&\textbf{62.8}&\underline{\textbf{\textcolor{red}{52.7}}}&\underline{\textbf{16.7}}&\underline{\textbf{28.3}}&\underline{\textbf{61.1}}&\underline{\textbf{56.6}}&\textbf{58.0}&\underline{\textbf{18.5}}&\underline{\textbf{47.8}}&\underline{\textbf{5.1}}&\textbf{36.3}&\underline{\textbf{53.3}}&\textbf{66.8}&\textbf{6.8}&\textbf{24.2}&\underline{\textbf{47.1}}&\textbf{11.0}&\underline{\textbf{\textcolor{red}{60.1}}}&\underline{\textbf{43.4}}&\underline{\textbf{41.3}} \\ 
\hline
OICR~\cite{tang2017multiple}& 71.0&59.1&42.3&27.4&20.2&58.7&46.4&18.6&18.1&45.7&21.7&20.5&53.1&68.5&1.8&15.7&42.7&40.0&41.0&61.5&38.7\\ 
OICR+ADR &67.0&63.1&50.8&12.8&23.8&55.3&55.1&16.1&5.2&47.2&23.4&28.2&55.9&69.2&1.9&21.5&46.5&49.9&35.9&63.8&39.6\\
\textbf{C-WSL:OICR+ADR}&\underline{\textbf{71.3}}&\underline{\textbf{68.3}}&\underline{\textbf{50.9}}&\textbf{17.1}&\underline{\textbf{24.8}}&\underline{\textbf{60.9}}&\underline{\textbf{56.4}}&\textbf{13.9}&\textbf{14.5}&\underline{\textbf{54.6}}&\textbf{22.2}&\textbf{25.7}&\underline{\textbf{57.7}}&\underline{\textbf{70.4}}&\textbf{1.6}&\textbf{20.0}&\underline{\textbf{\textcolor{red}{55.8}}}&\textbf{46.0}&\textbf{35.7}&\textbf{62.9}&\underline{\textbf{41.5}}  \\ 
 \hline
\textbf{C-WSL:ODR} &\textbf{74.0}&\textbf{67.3}&\textbf{45.6}&\textbf{29.2}&\textbf{26.8}&\textbf{62.5}&\textbf{54.8}&\textbf{21.5}&\textbf{22.6}&\textbf{50.6}&\textbf{24.7}&\textbf{25.6}&\textbf{57.4}&\textbf{71.0}&\textbf{2.4}&\textbf{22.8}&\textbf{44.5}&\textbf{44.2}&\textbf{45.2}&\textbf{\textcolor{red}{66.9}}&\textbf{43.0}  \\ 
\textbf{C-WSL:ODR+FRCNN} &\textbf{\textcolor{red}{75.3}}&\textbf{\textcolor{red}{71.6}}&\textbf{52.6}&\textbf{\textcolor{red}{32.5}}&\textbf{\textcolor{red}{29.9}}&\textbf{\textcolor{red}{62.9}}&\textbf{56.9}&\textbf{16.9}&\textbf{\textcolor{red}{24.5}}&\textbf{\textcolor{red}{59.0}}&\textbf{28.9}&\textbf{27.6}&\textbf{\textcolor{red}{65.4}}&\textbf{\textcolor{red}{72.6}}&\textbf{1.4}&\textbf{23.0}&\textbf{49.4}&\textbf{52.3}&\textbf{42.4}&\textbf{62.2}&\textbf{\textcolor{red}{45.4}} \\\hline
\end{tabular}
}
\label{tab: map_2012}

\end{table*}

\begin{table*}
\footnotesize
\setlength{\tabcolsep}{2pt}
\centering
\caption{Comparison with the state-of-the-art in terms of \emph{CorLoc} on the VOC2012 \emph{train} set. Our number is marked in \textcolor{red}{red} if it is the best in the column. \underline{Underline} is used if the C-WSL variant outperforms its baselines}
\resizebox{\textwidth}{!}{%
\begin{tabular}{l|cccccccccccccccccccc|c}
\hline
Methods & are & bik & brd & boa & btl & bus & car & cat & cha & cow & tbl & dog & hrs & mbk & prs & plt & shp & sfa & trn & tv & Avg. \\ \hline
OICR-Ens.+FRCNN~\cite{tang2017multiple}&85.4&81.5&70.4&44.7&46.6&83.6&78.4&33.9&29.3&83.2&51.6&50.5&86.1&88.0&11.0&56.7&82.5&\textcolor{black}{69.1}&65.1&83.6&64.1 \\
\hline
WSLPDA~\cite{Li_2016_CVPR} &80.5&63.7&64.4&34.1&29.3&76.7&71.5&62.8&30.3&76.1&23.0&55.3&75.2&77.7&18.7&56.4&66.7&25.1&66.5&54.8&55.4 \\ 
WSLPDA+ADR &87.2&79.7&72.4&38.6&40.9&82.6&75.2&\textcolor{black}{79.8}&35.1&81.3&18.9&\textcolor{black}{62.1}&82.4&83.9&\textcolor{black}{21.6}&\textcolor{black}{60.9}&75.4&29.5&74.5&55.5&61.9\\ 
\textbf{C-WSL:WSLPDA+ADR}&\textbf{85.7}&\textbf{77.2}&\underline{\textbf{73.4}}&\underline{\textbf{38.6}}&\underline{\textbf{46.4}}&\underline{\textbf{84.9}}&\underline{\textbf{75.8}}&\textbf{69.1}&\underline{\textbf{43.0}}&\textbf{76.8}&\textbf{20.1}&\textbf{58.6}&\textbf{79.8}&\textbf{79.6}&\textbf{20.3}&\textbf{57.8}&\underline{\textbf{79.5}}&\underline{\textbf{35.4}}&\underline{\textcolor{red}{\textbf{76.4}}}&\underline{\textbf{61.9}}&\underline{\textbf{62.0}} \\ 
\hline
OICR~\cite{tang2017multiple}&86.6&80.4&65.2&57.6&42.1&85.4&72.5&28.0&45.7&79.4&46.2&34.0&78.2&87.2&7.5&55.0&83.6&58.5&62.2&84.3&62.0\\ 
OICR+ADR &84.5&79.0&72.4&39.0&47.1&83.6&79.9&31.9&25.0&84.5&48.7&48.3&87.8&88.7&13.3&55.0&82.5&67.4&65.1&83.9&63.4\\
\textbf{C-WSL:OICR+ADR}&\underline{\textbf{86.6}}&\underline{\textbf{80.8}}&\underline{\textcolor{red}{\textbf{73.9}}}&\textbf{43.2}&\textbf{44.4}&\underline{\textcolor{red}{\textbf{87.7}}}&\textbf{76.2}&\underline{\textbf{32.2}}&\textbf{34.0}&\underline{\textcolor{red}{\textbf{87.1}}}&\underline{\textbf{49.1}}&\textbf{46.2}&\underline{\textcolor{red}{\textbf{88.2}}}&\underline{\textcolor{red}{\textbf{91.2}}}&\textbf{12.1}&\underline{\textbf{57.1}}&\textbf{78.4}&\textbf{65.5}&\underline{\textbf{65.1}}&\underline{\textbf{85.3}}&\underline{\textbf{64.2}}  \\ 
 \hline
\textbf{C-WSL:ODR} &\textbf{90.9}&\textbf{81.1}&\textbf{64.9}&\textbf{57.6}&\textbf{50.6}&\textbf{84.9}&\textbf{78.1}&\textbf{29.8}&\textbf{49.7}&\textbf{83.9}&\textbf{50.9}&\textbf{42.6}&\textbf{78.6}&\textbf{87.6}&\textbf{10.4}&\textbf{58.1}&\textbf{85.4}&\textbf{61.0}&\textbf{64.7}&\textcolor{red}{\textbf{86.6}}&\textbf{64.9}\\ 
\textbf{C-WSL:ODR+FRCNN} &\textcolor{red}{\textbf{92.1}}&\textcolor{red}{\textbf{84.3}}&\textbf{69.9}&\textcolor{red}{\textbf{58.3}}&\textcolor{red}{\textbf{53.9}}&\textbf{86.8}&\textcolor{red}{\textbf{80.4}}&\textbf{30.6}&\textcolor{red}{\textbf{52.6}}&\textbf{83.9}&\textcolor{red}{\textbf{54.7}}&\textbf{45.8}&\textbf{83.2}&\textbf{90.1}&\textbf{12.7}&\textbf{56.4}&\textcolor{red}{\textbf{86.0}}&\textbf{64.9}&\textbf{66.5}&\textbf{84.3}&\textcolor{red}{\textbf{66.9}} \\\hline
\end{tabular}
}
\label{tab: corloc_2012}
\end{table*}

\begin{figure}
\centering
   \includegraphics[width=1.0\linewidth]{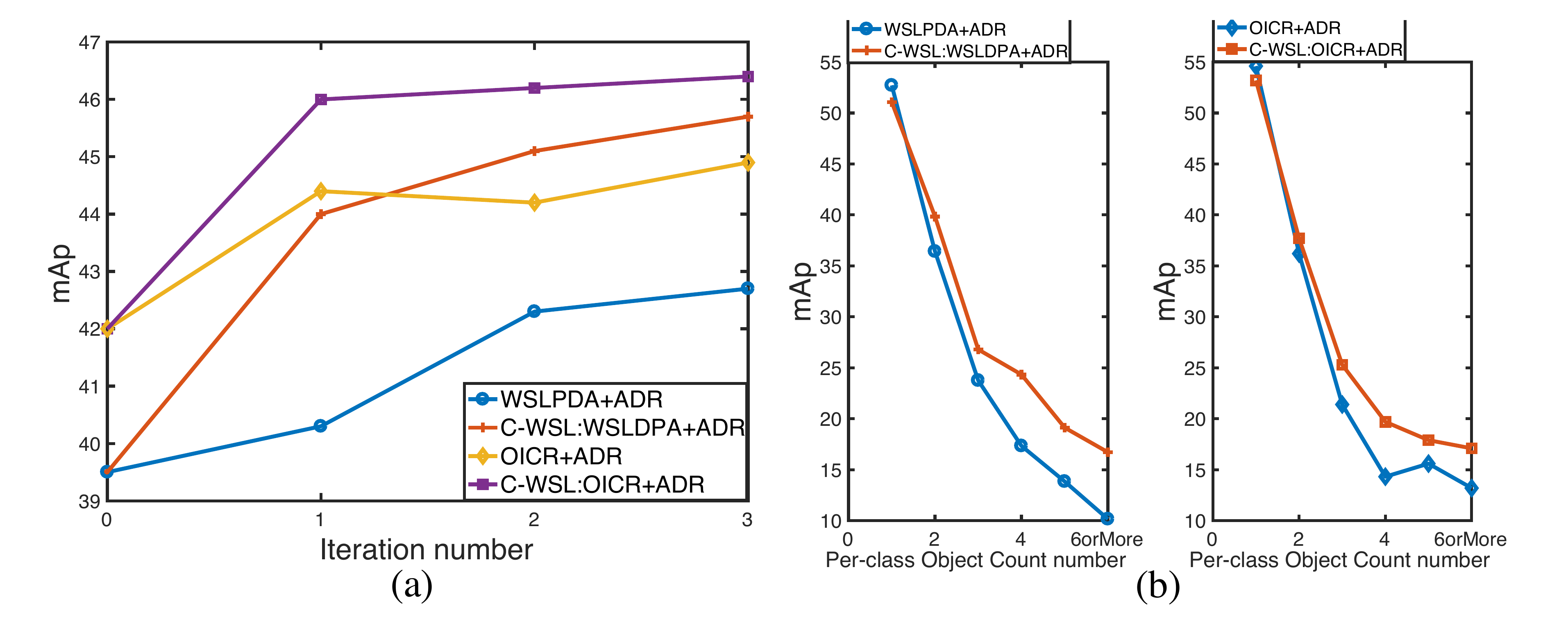}
   \caption{(a): model improvement as the number of \textit{ADR} iterations increases on the VOC2007 \emph{test} set. \textit{C-WSL} approaches improve faster than others. (b): Evaluation on images with different per-class object counts on VOC2007. Our approach outperforms the WSL detectors in the presence of multiple instances in a test image}
\label{fig: ablation}
\end{figure}

We also evaluated our methods and baselines (pre-trained on the VOC2007 trainval set) on the common 20 classes in MS COCO~\cite{lin2014microsoft} 35k-val2014 set using COCO mAP@0.5 metric. Although not fine-tuned on COCO, our approaches still outperform the baseline methods. The results are that C-WSL:WSLPDA improves WSLPDA~\cite{Li_2016_CVPR} from 17.9$\%$ to 19.6$\%$. C-WSL:OICR+ADR improves OICR~\cite{tang2017multiple} from 18.7$\%$ to 20.1$\%$ and C-WSL:ODR+FRCNN improves OICR-Ens.+FRCNN~\cite{tang2017multiple} from 19.0$\%$ to 20.0$\%$.

\subsection{Ablation Analysis}
Two major components contribute to the success of our approach. One is the iterative training process (alternating/online) and the other one is the per-class object count supervision. In Tab.~\ref{tab: adr_map} and~\ref{tab: adr_corloc}, we can see the improvement by adding ADR and object count into the system. For WSLPDA~\cite{Li_2016_CVPR}, iterative training (ADR) improves \emph{mAP} by $3.2\%$ and the count information (CRS) increases it by $3\%$. For OICR~\cite{tang2017multiple}, ADR helps by increasing $3.7\%$ \emph{mAP} and CRS contributes $1.5\%$. In the following, we analyze each component in detail.

\textit{Number of iterations.} 
\label{sec: abl-iters}
ADR performances as a function of the number of iterations using the WSLDPA and OICR models is shown in Fig.~\ref{fig: ablation}(a). Generally, models improve as the number of iterations increases. When adding object count supervision into the framework, the results of both WSLDPA and OICR models improve faster, which demonstrates the advantage of count information in WSL.

\textit{Number of object instances per image.}
\label{sec: abl-counts}
Adding the object count constraint helps a detector focus on a single object rather than multiple objects. To demonstrate this, we partition images in the VOC2007 \emph{test} set based on their per-class object count and re-evaluate our approaches on each subset.

The results are shown in Fig.~\ref{fig: ablation}(b). For both WSLPDA and OICR, the performance is much better under C-WSL. Generally, the gaps between curves of with and without C-WSL are bigger as the object count number increases.
\subsection{Error Analysis}
\begin{figure}[t]
\centering
   \includegraphics[width=1.0\linewidth]{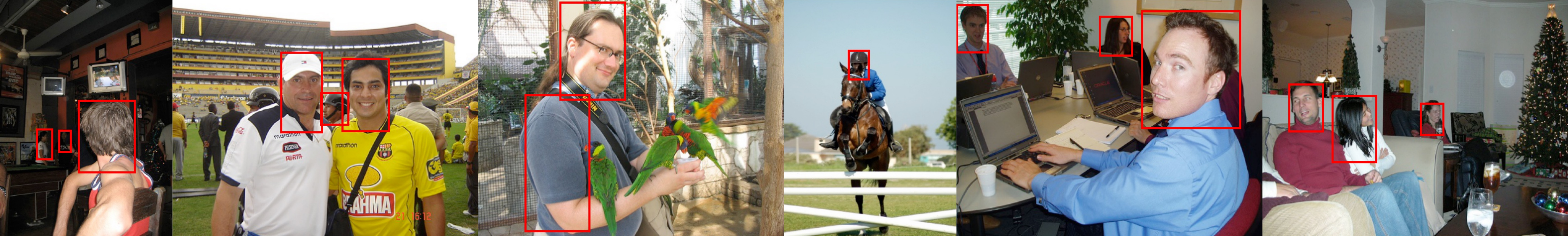}
   \caption{Some examples of the common failure cases of our approach (\emph{C-WSL:ODR+FRCNN}) on the ``person" category of the VOC2007 \emph{test} set}
\label{fig: failure_person}
\end{figure}

The results shown in Tab.~\ref{tab: map},~\ref{tab: corloc},~\ref{tab: map_2012} and~\ref{tab: corloc_2012} suggest that most existing WSL detectors perform poorly on the ``person" category: strongly supervised detectors achieve more than $76\%$ AP on the VOC2007 \emph{test} set (\emph{e.g.}, $76.6\%$~\cite{liu2016ssd} and $76.3\%$~\cite{ren2015faster}), while the best WSL detection result on ``person" is $20.3\%$ (see Tab.~\ref{tab: map}). This result is likely due to the large appearance variations of persons in the dataset. Without constraints provided by tight bounding boxes, rigid parts are easier to learn and mostly sufficient to differentiate the object from others. So, WSL detectors focus on local parts instead of the whole object as shown in Fig.~\ref{fig: failure_person}.

Intuitively, this can be overcome if we can roughly estimate the size of object instances. We conducted a preliminary experiment as follows. Suppose that we know the size of the smallest instance of an object category in an image and assume all the object parts are smaller than the smallest object. This assumption is not generally true and we use it just as a proof-of-concept. We preprocess the region candidates by removing all boxes whose size is smaller than the smallest object and then conduct \emph{C-WSL:WSLPDA+ADR} on VOC2007. The \emph{AP} on ``person" improves to $40.0\%$ and the \emph{mAP} over all the classes improves to $52.7\%$.

\section{Conclusions}
We proposed a Count-guided Weakly Supervised Localization (C-WSL) framework where a cheap and effective form of image-level supervision, \emph{i.e.}, per-class object count, is used to select training regions each of which tightly covers a single object instance for detector refinement. As a part of C-WSL, we proposed a Count-based Region Selection (CRS) algorithm to perform high-quality region selection. We integrated CRS into two detector refinement architectures to improve WSL detectors. Experimental results demonstrate the effectiveness of C-WSL. To prove the inexpensiveness of the per-class object count annotation, we conduct annotation experiments on VOC2007. The results show that only a small amount of time is needed to obtain the count information in an image and that we reduce the annotation time of center click and bounding box by more than $2\times$ and $38\times$ respectively.
\\ 
\textbf{Acknowledgement.} The research was supported by the Office of Naval Research under Grant N000141612713: Visual Common Sense Reasoning for Multi-agent Activity Prediction and Recognition. The authors would like to thank Eddie Kessler for proofreading the manuscript. 
%

\bibliographystyle{splncs04}
 \bibliography{egbib}

\end{document}